# SceneCalib: Automatic Targetless Calibration of Cameras and Lidars in Autonomous Driving

Ayon Sen, Gang Pan, Anton Mitrokhin, Ashraful Islam

*Abstract*— Accurate camera-to-lidar calibration is a requirement for sensor data fusion in many 3D perception tasks. In this paper, we present SceneCalib, a novel method for simultaneous self-calibration of extrinsic and intrinsic parameters in a system containing multiple cameras and a lidar sensor. Existing methods typically require specially designed calibration targets and human operators, or they only attempt to solve for a subset of calibration parameters. We resolve these issues with a fully automatic method that requires no explicit correspondences between camera images and lidar point clouds, allowing for robustness to many outdoor environments. Furthermore, the full system is jointly calibrated with explicit cross-camera constraints to ensure that camera-to-camera and camera-to-lidar extrinsic parameters are consistent.

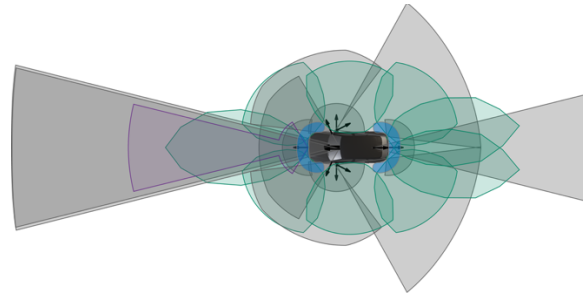

Figure 1. A typical autonomous vehicle sensor configuration (NVIDIA's Hyperion 8.1 platform). Camera FOVs are shown in gray, radar FOVs in green, and a front grille-mounted lidar in purple.

## I. INTRODUCTION

Autonomous vehicles typically incorporate several different sensor types for maximum coverage of their surroundings and robustness to different environmental conditions, as shown in Fig. 1. An array of cameras with various fields-of-view provides high resolution visual information, whereas lidar sensors provide direct measurements of depth at a sparser set of points around the vehicle. In order to provide a consistent description of the vehicle's surroundings, all sensors must be accurately registered to the same coordinate system, and intrinsic sensor properties must be adequately captured.

A common approach to solving this problem involves a *static calibration* process in which the vehicle is parked in a garage and views of a specially designed calibration target are captured [1], [2], [3], [4]. The target is designed with features that can be easily detected in each sensor, establishing correspondences used for calibration. Static calibration does not scale to a large fleet of vehicles due to the time and/or human effort involved. Periodic re-calibration may also be required if the sensors shift slightly between drives.

To address these issues, several approaches exist for performing calibration automatically using data collected from the environment that do not require specific targets or driving behavior. These methods typically require extracting specific types of features that can be reliably detected in both camera images and lidar point clouds to establish correspondence. Alternatively, they may involve maximizing correlation between signals in very different domains. Outdoor environments may have a significant amount of variation in terms of the landmarks available, so cross-sensor correspondence requirements can degrade the robustness of these approaches. Furthermore, many existing techniques only calibrate a single pair of sensors or assume camera intrinsic parameters are already known.

Our method, termed SceneCalib, does not require finding cross-modal correspondences and can jointly calibrate all extrinsic parameters and camera intrinsic parameters in a multi-camera/single-lidar system. This is achieved by:

- relying only on image correspondences without assuming *a priori* knowledge of which scene points they correspond to,
- demonstrating a reliable method for finding cross-camera image correspondences, and
- minimizing a purely geometric loss function between image feature pairs that constrains structure estimates to surfaces derived from lidar point clouds.

## II. RELATED WORK

Many approaches for performing targetless camera-to-lidar calibration (e.g., without targets specifically placed in the environment) exist in the literature, and they can be broadly categorized into two groups.

*Correspondence-based methods* seek to find calibration parameters that maximize alignment between features that have a detectable signal in both the camera images and the lidar point clouds. One common approach is to extract straight line or edge features from images and assume they must correspond to sharp discontinuities in the lidar depth. Levinson *et al* [5] and Kang *et al* [6] construct differentiable loss functions penalizing edge misalignment, and Cui *et al* [7] specifically explores line-based alignment for panoramic cameras. Ma *et al* [8] searches for lane edges and poles as a source of line features. Munoz-Banon *et al* [9] adds some preprocessing logic to extract object edges and associated direction vectors to create a signature that can be aligned

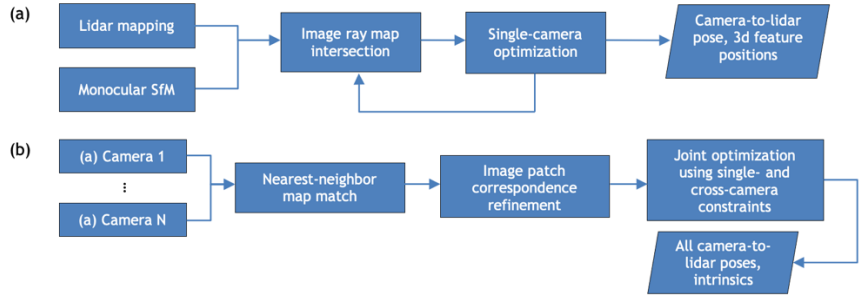

Figure 2. Overall flow of SceneCalib algorithm; (a) per-camera processing, (b) joint-camera processing.

between domains. Yuan *et al* [10] introduces a method to mitigate the effects of occlusion and bloom when extracting edges in lidar point clouds. Planar surfaces can also be used for establishing correspondence; Tamas *et al* [11] constructs an algebraic error that penalizes non-overlap in the camera image projection and is able to determine intrinsic parameters if two non-coplanar pairs are found (which are assumed to be known *a priori*).

Another correspondence-based method is direct comparison of photometric information from both sensors. Some lidar sensors collect intensity information along with depth, and changes in these values should be correlated to changes in pixel intensity across camera images. Pandey *et al* [12] presents a mutual information maximization method based on exactly this principle, and Taylor *et al* [13] adds lidar plane normals as another source of information. Shi *et al* [14] introduces an occlusion filtering method on top of the mutual information maximization approach. Other methods combine multiple types of correspondences to form a multi-objective optimization problem [15], [16].

End-to-end learning-based methods for calibration exist as well. The networks in these approaches directly estimate the amount of miscalibration error given camera-to-lidar extrinsics. These can be considered correspondence-based methods since the network contains layers for explicitly extracting correspondences between the lidar data and camera data [17], [18].

Finally, some methods have been developed for extracting correspondences based on high-level object detection. Semantic segmentation of the scene in both the image domain and lidar domain is performed to find where specific object types appear, allowing correspondences to be established. Nagy *et al* [19], [20] and Yoon *et al* [21] extract objects from a structure-from-motion point cloud derived from only camera frames to align to the same objects extracted from the lidar point cloud. Liu *et al* [22] performs the alignment between the lidar object projections and the segmented camera image pixels.

Because correspondence-based methods attempt to find features in two very different sensing modalities, it can be challenging to obtain accurate correspondences reliably, and it is a relatively strict requirement that a feature can be detected in both modalities. While straight line features may be prevalent in many environments, there is no guarantee that the same edges appear in both sensors. Objects in the lidar point cloud can look larger than they really are due to bloom, so the depth discontinuity will be offset from the edge in image space [10]; occlusion due to different sensor mounting positions also means the same features may not be visible in all sensors, and therefore occlusion handling is required. Photometric matching across sensor domains requires the presence of lidar intensity information that is not always available or accurate. Object-based correspondence is the most restrictive approach, requiring the use of pre-trained detectors for known object classes in both images and point clouds, so objects that the detectors have not been trained upon cannot be used.

To avoid correspondence detection altogether, some *correspondence-free methods* have been developed. One method of registration between the camera and lidar coordinate systems is to use an odometry trajectory derived independently from both sensors and determine coarse alignment by solving the hand-eye calibration problem. Typical structure-from-motion algorithms are used to solve for camera odometry (up to scale), and ICP is used to solve for lidar odometry. Taylor *et al* [23] uses per-sensor odometry as the primary constraint across each pair of sensors (allowing for camera-to-lidar and camera-to-camera constraints), but they combine this approach with photometric correspondence for finer alignment. Park *et al* [24] solves a bundle adjustment problem where the camera-to-lidar pose offset is constrained by the odometry alignment. On the other hand, Hu *et al* [25] iteratively solves for camera-to-lidar alignment with a geometric error and camera-to-camera alignment with photometric error for a stereo pair. The geometric error comes from misalignment between the stereo camera point cloud and the lidar point cloud, and the two objectives are alternately and iteratively refined until convergence.

While these methods avoid explicit cross-sensor correspondence, they do not allow all sensor measurements to be used simultaneously to constrain all calibration parameters. In addition, no existing methods solve for the extrinsic and intrinsic parameters for generic multi-camera configurations without requiring explicit camera-to-lidar feature correspondence.

III. METHODS

We propose a method, illustrated in Fig. 2, that relies only upon image feature correspondences between camera frames and the relatively weak assumption that image features are locally planar. We construct an optimization problem that minimizes a geometric loss function that encodes the notion that corresponding image features are views of the same point on a locally planar surface (surfel or mesh) reconstructed from lidar scans, essentially removing structure estimation from the optimization problem.

*A. Preliminaries*

We assume that the system consists of $N$ cameras with different intrinsic parameters and one lidar sensor. We require an invertible camera model for backprojection of image pixels as rays.

*Monocular structure-from-motion*: A monocular SfM method, such as ORB-SLAM [26], is applied to images from each camera to extract feature tracks – lists of pixels across multiple frames that correspond to the same world point. We want to emphasize that monocular SfM is used for extracting high-quality 2D feature tracks only, and we do not use its 3D structure or motion results.

*Lidar mapping*: Lidar scans provide a 3D point cloud, usually at a lower frame rate than cameras. At the lidar mapping step, we accumulate the lidar scans after ICP alignment (initialized by an IMU-based egomotion estimate). Moving objects are removed by checking for outliers when aligning scans. From the accumulated lidar scans, we build a ground surface mesh using Poisson Surface Reconstruction [27] and surfels [28] for structures above the ground. The ground mesh and structure surfels together provide the locally planar surface representation of the scene.

*B. Geometric Loss Function*

We define a geometric loss function based on two-view feature matching and known structures in the lidar map (see Figure 3). Let $u_{i,t_1}$ be an pixel in the $i$th camera frame at timestamp $t_1$, and let $u_{j,t_2}$ be a pixel in the $j$th camera's frame at timestamp $t_2$. If $u_{i,t_1}$ and $u_{j,t_2}$ are different views of the same point in the scene, their backprojected image rays should intersect at that point on a locally planar surface. In the case where there is misalignment, the spread of ray-to-plane intersection points quantifies the degree of misalignment. This metric can be expressed in image space to control for the effect of distance.

Let $\Pi_i$ represent the function that projects a camera-centered 3D point to its image pixel for the $i$th camera; this is a function only of the camera's intrinsic parameters. Let $T_i^L \in SE(3)$ represent the pose that transforms the camera-centered coordinate system for the $i$th camera to the lidar-centered coordinate system $L$; similarly, let $T_L^W(t) \in SE(3)$ represent the pose that transforms $L$ at timestamp $t$ into the map coordinate system $W$. Finally, let the normal and position $(n^W, p^W)$ parameterize a specific plane in the map coordinate system.

The operation $\Gamma$ transforms the image ray into map coordinates via $T_L^W(t_1)T_i^L$ and finds the point of intersection $x^W$ with the plane $(n^W, p^W)$. Then:

$$x^W = \Gamma\big(\Pi_i^{-1}(u_{i,t_1}), T_L^W(t_1)T_i^L, n^W, p^W\big) \quad (1)$$

$$e_{i,j,t_1,t_2} = \big\|\Pi_j\big((T_L^W(t_2)T_j^L)^{-1}x^W\big) - u_{j,t_2}\big\|^2 \quad (2)$$

$$e = \sum_{i,j,t_1,t_2} e_{i,j,t_1,t_2} \quad (3)$$

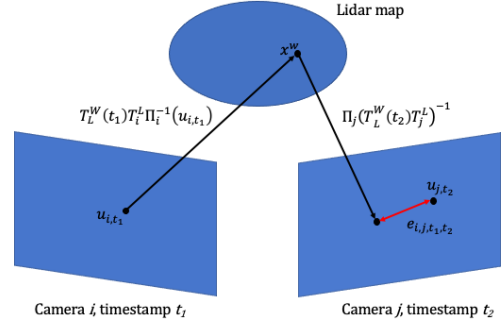

Figure 3. Geometric loss function minimized during automatic calibration. The 3D point $x^W$ is obtained by intersecting the back-projected ray with locally planar surfaces in the lidar map. The loss characterizes the difference between the feature transferred from another frame and the independent observation of the feature in the current frame.

The unknowns are the $N$ camera-to-lidar poses and camera intrinsic parameters $\{(T_i^L, \Pi_i)\}_{i=1}^N$. Note that $T_L^W(t)$ is known from the lidar trajectory obtained in the lidar mapping step.

Since we assume that the association of the plane with the image ray is known *a priori*, the operation $\Gamma$ is differentiable. This formulation allows refinement of the unknowns while constraining the structure points to the mesh surface, but without prescribing their 3D positions. It also makes no distinction between whether the pixel correspondences come from the same camera (e.g., $i = j$) or not.

*C. Track-to-Map Association*

Since the plane normals for the features must be known in the loss function, they are derived from the planar surface element that the backprojected image rays intersect with. This process is performed using initial estimates of the camera-to-lidar poses and camera intrinsics, so the accuracy of this association depends on how poor those initial estimates are. However, it also depends on how large the planar surface elements are. In a typical outdoor scene, there are features of many scales – the road surface, for example, may be a very coarse mesh, but foliage on trees can manifest as a collection of very small surfels with very different normal directions. To utilize the information present at multiple scales, we progressively build a set of tracks associated with large-to-small feature scales while performing a coarse-to-fine refinement of the calibration parameters and intersection points. Specific details are given in Appendix A.

*D. Per-Camera Optimization*

We first implement a single-camera-to-lidar extrinsics optimization loop to improve upon the initial estimates, reject outlier tracks, and estimate feature point positions for use in cross-camera correspondence detection. At each iteration, track-to-map association is performed, low-score tracks are removed, and RANSAC is used to find an inlier set of tracks (using the per-track RMSE of (2)) and corresponding extrinsic parameters. Track-to-map association thresholds are made tighter, and the process is repeated, adding the new tracks to the inlier set until the extrinsic parameters have converged.

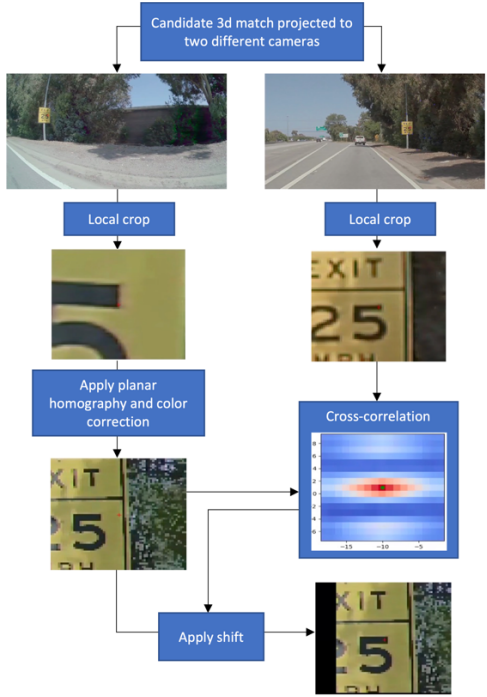

Figure 4. Cross-camera correspondence detection. Feature tracks from two cameras are associated via the lidar map. One viewpoint is warped to the other using an approximate planar homography and a more precise subpixel match is found using cross-correlation.

*E. Cross-Camera Correspondences*

Cross-camera correspondences are obtained in two steps: cross-camera track association, and image space match refinement. The processing flow is illustrated in Fig. 4. The result is a list of pixel correspondences between pairs of camera images from different cameras that were derived independently from the initially estimated camera-to-lidar poses (due to the image-space alignment). Note that this cross-camera feature matching is done without requiring simultaneously overlapping views.

*Cross-camera track association*: Via backprojection of feature tracks, the previous per-camera optimization step yields a set of $M_i$ feature points $\{x_k^W\}_{k=1}^{M_i}$ in the map coordinate system for each of the $i = 1, \ldots, N$ cameras. A nearest-neighbor search (limited to a small radius) between the point clouds for camera pair $(i, j)$ allows us to find which pairs of feature tracks (one from the $i$th camera, and one from the $j$th camera) are likely to be views of the same feature point.

*Image space match refinement*: Consider a single map point $x^W$ derived from the nearest-neighbor search. Let the pixel $u_{i,t_1}$ be the projection of $x^W$ into the $i$th camera's image at timestamp $t_1$ (taken from the feature track for the $i$th camera). Following the assumption that map features are locally planar, we assume that a small neighborhood of pixels $S_{i,t_1}$ centered upon $u_{i,t_1}$ (e.g., a 65x65 pixel patch) is an image of a plane. Then

$$H_{i,j,t_1,t_2} = \Pi_j (T_L^W(t_2) T_j^L)^{-1} T_L^W(t_1) T_i^L \Pi_i^{-1} \quad (4)$$

defines a planar homography. The values for $T_i^L, T_j^L, \Pi_i$, and $\Pi_j$ come from the individual single-camera-to-lidar optimization problems solved earlier, so this homography is not exact but can be used to correct most of the planar distortion. The last step is using simple cross-correlation between the image patches $H_{i,j,t_1,t_2}(S_{i,t_1})$ and $S_{j,t_2}$ to find a more precise pixel correspondence. In practice, this involves converting the patches to grayscale, equalizing their intensity histograms, and computing a subpixel-accurate alignment by finding the center of mass of the cross-correlation values.

*F. Joint Optimization*

Using the single- and cross-camera correspondences, we can now minimize the geometric loss function given in equation (3) with constraints between all sensor pairs. Cross-camera matches with wider cross-correlation peaks are removed, and RANSAC is used to determine which correspondences are inliers. Some cameras naturally will produce many more feature matches than others due to their placement and FOV, so the total sum of all residuals involving projection into a given camera is normalized to 1. Both camera-to-lidar extrinsic parameters and camera intrinsic parameters are optimized. Ceres Solver is used to implement the loss function and perform nonlinear least-squares optimization [29].

IV. RESULTS

We evaluate the performance of the algorithm using datasets recorded from vehicles with 12 cameras of varying fields of view (as shown in Fig. 1) and one 360-degree lidar placed on the roof of the vehicle. Camera intrinsic parameters are modeled using the f-theta model (see Appendix B). The eight non-fisheye cameras (30°, 70°, and 120° FOVs) record 3848x2168 resolution images at 30 fps, and the four fisheye cameras (200° FOV) record 1280x720 resolution images at 30 fps. Lidar spins are taken at 10 fps. Recordings are generally taken from daytime driving scenarios, and camera recordings are not synchronized with each other or with the lidar recordings.

Note that we only use a fraction of each recording, employing an automatic selection algorithm that scans the egomotion trajectory, extracting around five 1-km segments with significant motion. This still results in anywhere from 200 thousand to 2 million SfM features being used in the final joint optimization problem.

Several examples qualitatively illustrating the lidar-to-camera calibration accuracy achieved by SceneCalib are shown in Fig. 5.

*A. Miscalibration Analysis*

To quantify calibration quality, we develop a metric that is based off Levinson *et al*'s [4] miscalibration detection algorithm. For a pair of frames from the $i$th and $j$th cameras taken at similar timestamps, we project the lidar map on to both camera frames and compute the photometric error between projections of the same lidar point. Performing this process for many noisy values of $T_j^{L^{-1}}$, we count the number of samples with lower photometric error than the calibrated pose. We define the ratio of samples with lower error, averaged across many camera frame pairs, as the *miscalibration rate*. A perfectly calibrated, noise-free system should always have minimum photometric error when all deviations are zero, so this scenario is represented by a miscalibration rate of zero.

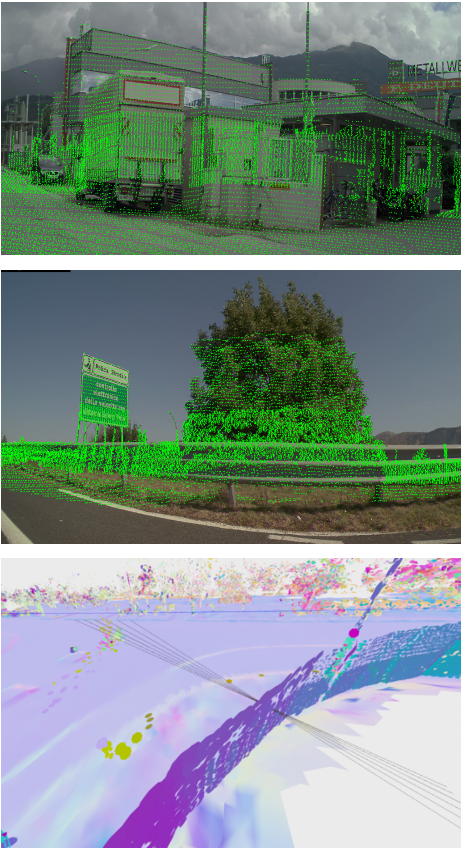

Figure 5. Lidar point cloud projections (in green) using SceneCalib results exhibit precise alignment with camera image features. Examples are shown for 70° (top) and 120° (middle) FOV cameras. After calibration, backprojected image rays converge to a point on a planar surface in the lidar map (bottom).

Pose perturbations are performed separately for each axis; position component ($x, y, z$) perturbations are uniformly randomly sampled from $[-2, 2]$ cm, and rotation component (roll, pitch, yaw) perturbations are uniformly randomly sampled from $[-0.2, 0.2]°$. Because the deviations are applied in the camera-centered coordinate system, they are reported in right-down-forward (RDF) coordinates.

### B. Comparison to Static Calibration

We can first compare the performance of SceneCalib to a static garage calibration process performed before a drive. The garage calibration consists of collecting views of a checkerboard target to solve for camera intrinsic parameters and camera-to-lidar extrinsic parameters [30]. The dataset used for comparison consists of 25 recordings from the same vehicle, ranging from 25 to 65 minutes in length and spanning a two-month period. For a fair comparison, SceneCalib is initialized using a result that is significantly offset from both the final result and the static calibration (e.g., by up to 0.5° in orientation and 5 cm in position).

As shown in Fig. 6, our automatic calibration algorithm has a similar or better miscalibration rate for all degrees of freedom when compared to static calibration. Since certain camera subsystems are used for different autonomous driving tasks, we divide the error based on the FOVs of the camera pairs. Note that wider FOV and lower resolution cameras have lower overall miscalibration rates since the metric is a measure of sensitivity and depends on the size of the interval the perturbations are sampled from.

### C. Robustness Across Multiple Driving Scenarios

Since static calibration is a time-consuming and manual process, it is hard to obtain a large dataset with a high level of scene diversity. Because of this, all 25 recordings in the previous section were taken from the same city. To determine whether our automatic calibration algorithm is robust across more diverse scenes, we compute miscalibration rates for only the automatic calibration results using a dataset of 55 recordings from several vehicles that are 30 to 70 minutes in length and span a 7-month period.

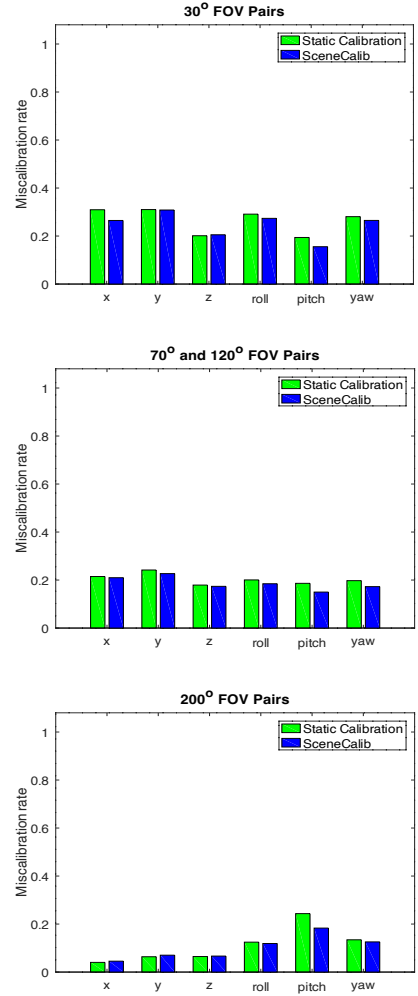

Figure 6. Comparison of miscalibration rates for static calibration and automatic calibration (SceneCalib) per axis, grouped by camera FOV.

These datasets were recorded in multiple cities from multiple countries during daytime hours.

Miscalibration rates for the more diverse dataset are comparable or only slightly worse in all camera categories and axes when compared to the results from the less diverse dataset (see Table 1).

### D. Weather and Lighting Conditions

While SceneCalib is robust to different scenes, some weather and lighting conditions are difficult to handle. In general, nighttime driving scenarios are unlikely to produce a good automatic calibration result due to the number of camera features produced by the SfM algorithm. While a daytime drive may produce hundreds of thousands of features, a nighttime drive may only produce hundreds or thousands. Other low-visibility conditions (such as heavy rain) are similarly difficult to deal with. Furthermore, recordings containing snow can be unsuitable for the lidar mapping step due to noisy lidar scans.

We typically avoid directly calibrating such recordings. Instead, we use a result from a different recording from the same vehicle taken under more favorable conditions, using the assumption that the extrinsic and intrinsic parameters will not change significantly over short periods of time.

TABLE I. ROBUSTNESS OF SCENECALIB TO INCREASED SCENE DIVERSITY

| FOV | Dataset | Miscalibration Rate | | | | | |
|---|---|---|---|---|---|---|---|
| | | x | y | z | roll | pitch | yaw |
| 30° | Non-diverse | 0.265 | 0.308 | 0.205 | 0.273 | 0.155 | 0.265 |
| | Diverse | 0.294 | 0.313 | 0.224 | 0.299 | 0.173 | 0.294 |
| 70°, 120° | Non-diverse | 0.210 | 0.227 | 0.174 | 0.185 | 0.150 | 0.173 |
| | Diverse | 0.219 | 0.224 | 0.186 | 0.195 | 0.179 | 0.186 |
| 200° | Non-diverse | 0.045 | 0.070 | 0.066 | 0.119 | 0.183 | 0.126 |
| | Diverse | 0.079 | 0.117 | 0.095 | 0.144 | 0.182 | 0.148 |

## V. CONCLUSION

We propose a fully automatic, targetless calibration algorithm for autonomous vehicles with multiple cameras of different characteristics and a lidar sensor. The algorithm can calibrate camera-to-lidar extrinsic parameters, camera intrinsic parameters, and explicitly constrains camera-to-camera pose transformations while also constraining structure estimation with the high-quality 3D information provided by lidar. We have demonstrated that the algorithm is completely free of human intervention (and thus highly scalable), achieves calibration quality comparable to manual calibration, and is robust to various scenes. While low light and poor weather conditions are difficult to calibrate directly, a major area of future work will be to explore the feature detection module component. Significant improvements to the robustness of the algorithm could potentially be made by having the ability to extract more image features under such conditions.

## APPENDIX

### A. Track-to-Map Association Thresholds

We construct two scores for the quality of the ray bundle association with a planar surface. Let $i, j$ index features in the same track (which has $N$ observations) and let $x_i$ and $n_i$ represent the mesh intersection point and mesh intersection normal, respectively.

$$s_{\theta,i} = \sum_{j \neq i} \exp\left(-\frac{1}{2}\left(\frac{\text{acos}(n_i \cdot n_j)}{\sigma_\theta}\right)^2\right) \quad (5)$$

$$s_\theta = \frac{1}{N-1} \max_i s_{\theta,i} \quad (6)$$

The first score, $s_\theta$, is larger if the angle between the normal vectors of the intersected planes is small.

$$i^* = \underset{i}{\text{argmax}}\, s_{\theta,i} \quad (7)$$

$$s_d = \frac{1}{N-1} \sum_{i \neq i^*} \exp\left(-\frac{1}{2}\left(\frac{n_i \cdot (x_i - x_{i^*})}{\sigma_d}\right)^2\right) \quad (8)$$

The second, $s_d$, is larger if the distance between the intersection points on the plane is small. The parameters $\sigma_\theta$ and $\sigma_d$ can be adjusted to allow for smaller or larger spreads in these statistics.

### B. F-theta Model

We use a *captured rays-based model* [31] for the camera projection with a high-order polynomial function of the ray angle to capture distortion [32]. Given a point $p = [p_x, p_y, p_z]$ in the camera-centered world coordinate system, we compute the projected pixel $u$ via:

$$\hat{p} = [p_x/p_z, p_y/p_z], \quad r = \|\hat{p}\|, \quad \theta = \tan^{-1} r \quad (9)$$
$$f(\theta) = \sum_{i=1}^{5} k_i \theta^i \quad (10)$$
$$u = \Pi(p) = (f(\theta)/r)\hat{p} + u_0 \quad (11)$$

The parameters are the polynomial coefficients $k_i$ and the optical center $u_0$, and $\theta$ is the angle of the ray with the optical axis.

Note that this model is not analytically invertible due to (10). Since we require an initial guess for the intrinsic parameters in our optimization scheme, we solve for an approximate inverse $f^{-1}(x) \approx \sum_{i=1}^{5} m_i x^i$. Rather than optimizing $\{k_i\}$ and $\{m_i\}$, we introduce a scale factor $s_f$:

$$f(s_f \theta) = \sum_{i=1}^{5} (s_f^i k_i)\theta^i \quad (12)$$

and leave the polynomial coefficients fixed from the initial guess. This factor can be applied analytically to the approximate inverse without introducing much additional error. In practice, the higher order coefficients tend to be several orders of magnitude smaller than the linear coefficient and contribute very little to the overall variation in the polynomial. Thus, in our formulation, intrinsic parameters are tuned by solving for $s_f$ and $u_0$.

## ACKNOWLEDGMENT

The authors would like to thank Cheng-Chieh Yang for assistance with monocular SfM processing, Chen Chen for providing the static calibration dataset, and Yuchen Deng for providing lidar trajectory benchmarks.

## References

[1] L. Zhou, Z. Li, and M. Kaess, "Automatic Extrinsic Calibration of a Camera and a 3D LiDAR Using Line and Plane Correspondences," in *2018 IEEE/RSJ International Conference on Intelligent Robots and Systems (IROS)*, Madrid, Oct. 2018, pp. 5562–5569. doi: 10.1109/IROS.2018.8593660.

[2] Z. Chai, Y. Sun, and Z. Xiong, "A Novel Method for LiDAR Camera Calibration by Plane Fitting," in *2018 IEEE/ASME International Conference on Advanced Intelligent Mechatronics (AIM)*, Auckland, New Zealand, Jul. 2018, pp. 286–291. doi: 10.1109/AIM.2018.8452339.

[3] E. Kim and S.-Y. Park, "Extrinsic Calibration between Camera and LiDAR Sensors by Matching Multiple 3D Planes," *Sensors*, vol. 20, no. 1, p. 52, Dec. 2019, doi: 10.3390/s20010052.

[4] P. Baker and Y. Aloimonos, "Complete calibration of a multi-camera network," in *Proceedings IEEE Workshop on Omnidirectional Vision (Cat. No.PR00704)*, Hilton Head Island, SC, USA, 2000, pp. 134–141. doi: 10.1109/OMNVIS.2000.853820.

[5] J. Levinson and S. Thrun, "Automatic Online Calibration of Cameras and Lasers," in *Robotics: Science and Systems IX*, Jun. 2013. doi: 10.15607/RSS.2013.IX.029.

[6] J. Kang and N. L. Doh, "Automatic targetless camera–LIDAR calibration by aligning edge with Gaussian mixture model," *J. Field Robot.*, vol. 37, no. 1, pp. 158–179, Jan. 2020, doi: 10.1002/rob.21893.

[7] T. Cui, S. Ji, J. Shan, J. Gong, and K. Liu, "Line-Based Registration of Panoramic Images and LiDAR Point Clouds for Mobile Mapping," *Sensors*, vol. 17, no. 12, p. 70, Dec. 2016, doi: 10.3390/s17010070.

[8] T. Ma, Z. Liu, G. Yan, and Y. Li, "CRLF: Automatic Calibration and Refinement based on Line Feature for LiDAR and Camera in Road Scenes." arXiv, Mar. 08, 2021. Accessed: Sep. 13, 2022. [Online]. Available: http://arxiv.org/abs/2103.04558

[9] M. A. Munoz-Banon, F. A. Candelas, and F. Torres, "Targetless Camera-LiDAR Calibration in Unstructured Environments," *IEEE Access*, vol. 8, pp. 143692–143705, 2020, doi: 10.1109/ACCESS.2020.3014121.

[10] C. Yuan, X. Liu, X. Hong, and F. Zhang, "Pixel-Level Extrinsic Self Calibration of High Resolution LiDAR and Camera in Targetless Environments," *IEEE Robot. Autom. Lett.*, vol. 6, no. 4, pp. 7517–7524, Oct. 2021, doi: 10.1109/LRA.2021.3098923.

[11] L. Tamas and Z. Kato, "Targetless Calibration of a Lidar - Perspective Camera Pair," in *2013 IEEE International Conference on Computer Vision Workshops*, Sydney, Australia, Dec. 2013, pp. 668–675. doi: 10.1109/ICCVW.2013.92.

[12] G. Pandey, J. McBride, S. Savarese, and R. Eustice, "Automatic Targetless Extrinsic Calibration of a 3D Lidar and Camera by Maximizing Mutual Information," *Proc. AAAI Conf. Artif. Intell.*, vol. 26, no. 1, pp. 2053–2059, Sep. 2021, doi: 10.1609/aaai.v26i1.8379.

[13] Z. Taylor and J. Nieto, "Automatic Calibration of Lidar and Camera Images using Normalized Mutual Information," p. 8.

[14] C. Shi, K. Huang, Q. Yu, J. Xiao, H. Lu, and C. Xie, "Extrinsic Calibration and Odometry for Camera-LiDAR Systems," *IEEE Access*, vol. 7, pp. 120106–120116, 2019, doi: 10.1109/ACCESS.2019.2937909.

[15] J. Jeong, Y. Cho, and A. Kim, "The Road is Enough! Extrinsic Calibration of Non-overlapping Stereo Camera and LiDAR using Road Information," *IEEE Robot. Autom. Lett.*, vol. 4, no. 3, pp. 2831–2838, Jul. 2019, doi: 10.1109/LRA.2019.2921648.

[16] K. Irie, M. Sugiyama, and M. Tomono, "Target-less camera-LiDAR extrinsic calibration using a bagged dependence estimator," in *2016 IEEE International Conference on Automation Science and Engineering (CASE)*, Fort Worth, TX, USA, Aug. 2016, pp. 1340–1347. doi: 10.1109/COASE.2016.7743564.

[17] X. Lv, B. Wang, Z. Dou, D. Ye, and S. Wang, "LCCNet: LiDAR and Camera Self-Calibration using Cost Volume Network," in *2021 IEEE/CVF Conference on Computer Vision and Pattern Recognition Workshops (CVPRW)*, Nashville, TN, USA, Jun. 2021, pp. 2888–2895. doi: 10.1109/CVPRW53098.2021.00324.

[18] S. Wu, A. Hadachi, D. Vivet, and Y. Prabhakar, "This Is the Way: Sensors Auto-Calibration Approach Based on Deep Learning for Self-Driving Cars," *IEEE Sens. J.*, vol. 21, no. 24, pp. 27779–27788, Dec. 2021, doi: 10.1109/JSEN.2021.3124788.

[19] B. Nagy, L. Kovacs, and C. Benedek, "Online Targetless End-to-End Camera-LIDAR Self-calibration," in *2019 16th International Conference on Machine Vision Applications (MVA)*, Tokyo, Japan, May 2019, pp. 1–6. doi: 10.23919/MVA.2019.8757887.

[20] B. Nagy, L. Kovacs, and C. Benedek, "SFM And Semantic Information Based Online Targetless Camera-LIDAR Self-Calibration," in *2019 IEEE International Conference on Image Processing (ICIP)*, Taipei, Taiwan, Sep. 2019, pp. 1317–1321. doi: 10.1109/ICIP.2019.8804299.

[21] B.-H. Yoon, H.-W. Jeong, and K.-S. Choi, "Targetless Multiple Camera-LiDAR Extrinsic Calibration using Object Pose Estimation," in *2021 IEEE International Conference on Robotics and Automation (ICRA)*, Xi'an, China, May 2021, pp. 13377–13383. doi: 10.1109/ICRA48506.2021.9560936.

[22] Z. Liu, H. Tang, S. Zhu, and S. Han, "SemAlign: Annotation-Free Camera-LiDAR Calibration with Semantic Alignment Loss," in *2021 IEEE/RSJ International Conference on Intelligent Robots and Systems (IROS)*, Prague, Czech Republic, Sep. 2021, pp. 8845–8851. doi: 10.1109/IROS51168.2021.9635964.

[23] Z. Taylor and J. Nieto, "Motion-Based Calibration of Multimodal Sensor Extrinsics and Timing Offset Estimation," *IEEE Trans. Robot.*, vol. 32, no. 5, pp. 1215–1229, Oct. 2016, doi: 10.1109/TRO.2016.2596771.

[24] C. Park, P. Moghadam, S. Kim, S. Sridharan, and C. Fookes, "Spatiotemporal Camera-LiDAR Calibration: A Targetless and Structureless Approach," *IEEE Robot. Autom. Lett.*, vol. 5, no. 2, pp. 1556–1563, Apr. 2020, doi: 10.1109/LRA.2020.2969164.

[25] H. Hu, F. Han, F. Bieder, J.-H. Pauls, and C. Stiller, "TEScalib: Targetless Extrinsic Self-Calibration of LiDAR and Stereo Camera for Automated Driving Vehicles with Uncertainty Analysis." arXiv, Feb. 28, 2022. Accessed: Sep. 13, 2022. [Online]. Available: http://arxiv.org/abs/2202.13847

[26] R. Mur-Artal, J. M. M. Montiel, and J. D. Tardos, "ORB-SLAM: A Versatile and Accurate Monocular SLAM System," *IEEE Trans. Robot.*, vol. 31, no. 5, pp. 1147–1163, Oct. 2015, doi: 10.1109/TRO.2015.2463671.

[27] M. Kazhdan, M. Bolitho, and H. Hoppe, "Poisson surface reconstruction," p. 10.

[28] J. Behley and C. Stachniss, "Efficient Surfel-Based SLAM using 3D Laser Range Data in Urban Environments," in *Robotics: Science and Systems XIV*, Jun. 2018. doi: 10.15607/RSS.2018.XIV.016.

[29] S. Agarwal, K. Mierle, and The Ceres Solver Team, "Ceres Solver." Mar. 2022. [Online]. Available: https://github.com/ceres-solver/ceres-solver

[30] M. D. Wheeler and L. Yang, "Lidar to camera calibration for generating high definition maps," US10531004B2, Jan. 07, 2020 Accessed: Sep. 13, 2022. [Online]. Available: https://patents.google.com/patent/US10531004B2/en

[31] J. Courbon, Y. Mezouar, L. Eckt, and P. Martinet, "A generic fisheye camera model for robotic applications," in *2007 IEEE/RSJ International Conference on Intelligent Robots and Systems*, San Diego, CA, Oct. 2007, pp. 1683–1688. doi: 10.1109/IROS.2007.4399233.

[32] D. Scaramuzza, A. Martinelli, and R. Siegwart, "A Flexible Technique for Accurate Omnidirectional Camera Calibration and Structure from Motion," in *Fourth IEEE International Conference on Computer Vision Systems (ICVS'06)*, New York, NY, USA, 2006, pp. 45–45. doi: 10.1109/ICVS.2006.3.